\documentclass[letterpaper, 10 pt, conference]{ieeeconf}  

\IEEEoverridecommandlockouts                              

\usepackage{graphics} 
\usepackage{epsfig} 
\usepackage{mathptmx} 
\usepackage{times} 
\usepackage{amsmath} 
\usepackage{amssymb}  

\usepackage{todonotes}
\usepackage{xcolor}
\usepackage{xspace}
\usepackage{svg}
\usepackage{booktabs} 
\usepackage{cite}
\usepackage{hyperref}

\usepackage{adjustbox}
\usepackage{algorithm}
\usepackage{algpseudocode}

\usepackage{float}
\usepackage{cuted}
\usepackage{subfigure}
\usepackage{subcaption}

\usepackage{balance}

\title{\LARGE \bf Lightweight Visual Reasoning for Socially-Aware Robots}

\author{Alessio Galatolo$^{1*}$, 
Ronald Cumbal$^{1*}$, 
Alexandros Rouchitsas$^{1}$,
Katie Winkle$^{1}$, \\
Didem G{\"u}rd{\"u}r Broo$^{1}$, and
Ginevra Castellano$^{1}$
\thanks{*Shared-first authorship}
\thanks{$^{1}$Department of Information Technology,
        Uppsala University, Sweden. Mail correspondence to: 
        {\tt\small alessio.galatolo@it.uu.se}}%
}

\begin{document}

\maketitle
\thispagestyle{empty}
\pagestyle{empty}

\begin{strip}
\vspace{-15mm}
\centering
\includegraphics[width=1.0\textwidth]{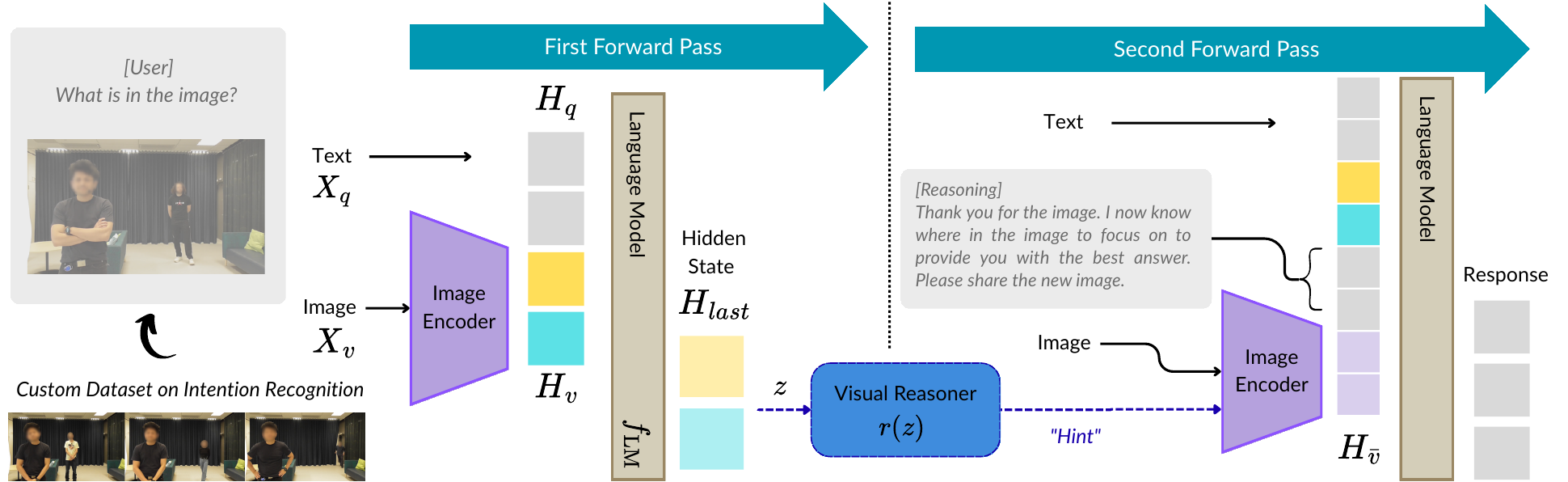}
    \captionof{figure}{\textbf{Overview of the Visual Reasoning approach}: A module connects an LLM's hidden states for image tokens back to the vision encoder through a gated MLP, creating a reasoning loop between text and vision. Training uses a two-pass strategy: the first pass extracts reasoning features from the LLM, and the second integrates them into the image encoding, reinterpreting the visual content in light of textual context and reasoning. Detailed description shown in Algorithm \ref{alg:training}. A sample of the custom dataset on intention recognition during human-robot interactions is shown in the bottom left.}
    \label{fig:complete_system}
\vspace{-10px}
\end{strip}

\begin{abstract}

Robots operating in shared human environments must not only navigate, interact, and detect their surroundings, they must also interpret and respond to dynamic, and often unpredictable, human behaviours. Although recent advances have shown promise in enhancing robotic perception and instruction-following using Vision-Language Models (VLMs), they remain limited in addressing the complexities of multimodal human-robot interactions (HRI). Motivated by this challenge, we introduce a lightweight language-to-vision feedback module that closes the loop between an LLM and the vision encoder in VLMs. The module projects image-token hidden states through a gated Multi-Layer Perceptron (MLP) back into the encoder input, prompting a second pass that reinterprets the scene under text context. We evaluate this approach on three robotics-centred tasks: navigation in a simulated environment (Habitat), sequential scene description (Mementos-Robotics), and human-intention recognition (our HRI dataset). Results show that our method improves Qwen 2.5 (7B) by $3.3\%$ (less distance), $+0.057$ description score, and $+2.93\%$ accuracy, with less than $3\%$ extra parameters; Gemma 3 (4B) and LLaVA OV 1.5 (4B) show mixed navigation results but gains $+0.111,+0.055$ and $+10.81\%,+4.79\%$ on the latter two tasks.
\end{abstract}

\section{Introduction}

Integrating robots into human-shared environments goes beyond completing tasks and ensuring physical safety. It also demands a deep understanding, and often anticipation, of dynamic contexts. However, reasoning about and responding to such environments is inherently challenging, particularly when human behaviour is involved. For instance, robots operating in urban spaces must not only perceive their surroundings to achieve socially aware navigation but also employ effective communication strategies \cite{singamaneni2024survey}. They are expected to react to spontaneous human behaviours \cite{weinberg2023sharing} and make context-dependent decisions about how to engage with nearby individuals \cite{sasabuchi2025agreeing, moujahid2022multi}. These challenges---navigating diverse environments, understanding their characteristics, and managing human interactions---highlight the critical role of accurate environment reasoning in the successful deployment of robots in human-shared spaces.

Recent advancements have shown that Large Language Models (LLMs) can enable robots to better interpret and follow human instructions \cite{kim2024survey}. Similarly, Vision-Language Models (VLMs) have been proposed as a way to connect advanced visual perception capabilities to text, allowing robots to receive natural language instructions and being able to reason about their surroundings. Despite these developments, much of the existing research addresses these capabilities in isolation, overlooking the holistic integration required for robots to operate effectively in human-shared environments. In particular, the role of Human-Robot Interaction (HRI) is often overlooked \cite{li2025benchmark}. To bridge this gap, our work examines how VLMs can be leveraged to strengthen robots' capacity to navigate diverse scenarios, with special attention to the challenges posed by complex human behaviours.

In particular, we assume that understanding human behaviour requires quite advanced reasoning and understanding of small cues \cite{tapus2019perceiving}. We thus look at approaches that integrate reflection and reasoning mechanisms into VLMs. While techniques such as Chain-of-Thought (CoT) prompting \cite{cot} and structured reasoning \cite{snell2024scalingllmtesttimecompute} have significantly improved the performance of LLMs, their application in the context of multimodal models remains underexplored. Existing approaches to multimodal reasoning often rely on shallow combinations of visual and textual inputs, using visual information merely as context for textual reasoning rather than achieving deep integration of the two modalities \cite{thawakar2025llamavo1rethinkingstepbystepvisual, zhang2024multimodal, zhang2024improvevisionlanguagemodel}. 
To address this limitation, we introduce a novel reasoning module that establishes a direct feedback loop between visual and textual modalities. This deeper integration enables more robust interpretation of complex environments.

Our contributions are twofold: 
\begin{itemize}
    \item We propose a lightweight visual reasoning module that enables the language model to modulate the vision encoder, closing the loop between perception and interpretation---an underexplored architectural principle in current VLMs for robotics.
    \item We show empirical gains on scene description and intention recognition, and modest gains on navigation, analysed via ablations on image reuse, MLP removal, and order of input modality.
\end{itemize}

\section{Related Works}

LLMs have become powerful tools in the field of robotics, offering the ability to interpret complex instructions, reason through tasks, and communicate more effectively with humans using natural language \cite{kim2024survey, ahn2022can, zeng2023demonstrating}. At the same time, the integration of multimodal inputs---especially visual data---has enhanced the capabilities of LLMs beyond text-based understanding \cite{liu2024llavanext, Liu2023Visual, zhu2023minigpt4, wang2024cogvlm}. 


\paragraph{Vision models in robotic systems}

Pre-trained VLMs, such as CLIP \cite{clip} and InstructBLIP \cite{li2023blip}, have played a pivotal role in enabling robots to process visual inputs for tasks such as object recognition and scene understanding \cite{kim2024survey}. For example, Kwon et al. \cite{kwon2024toward} proposed combining LLMs with VLMs to facilitate grounded commonsense reasoning, allowing robots to actively perceive and interpret their environment. Similarly, Sermanet et al. \cite{sermanet2024robovqa} introduced RoboVQA, which leverages video input to support decision-making and visual understanding in complex, real-world scenarios. 
Furthermore advancing this line of work, Li et al. \cite{li2024mmro} introduced MMRo, a benchmark designed to evaluate robotic skills such as spatial reasoning, task planning, and safety awareness. Their findings indicate that even state-of-the-art models still face challenges in basic perceptual tasks, such as accurately identifying object attributes like colour, shape, material, and spatial location. 

While these models have improved object and scene understanding in robotics, they largely overlook the interpretation of human behaviours, goals, or intentions ---an ability considered fundamental to intelligent, cooperative systems \cite{dennett1989intentional, belardinelli2024gaze}. Research in HRI has attempted to address this challenge by studying human engagement \cite{glas2015definitions}
, turn-taking \cite{skantze2021turn}, and interactions involving multiple participants \cite{moujahid2022multi}. However, the few efforts that apply vision models in these contexts show that even state-of-the-art systems continue to struggle in open-ended scenarios \cite{sasabuchi2025agreeing}. 
To address this gap, our work focuses on developing a reasoning module specifically designed to enhance the robot’s ability to interpret to human behaviour in complex, real-world settings.

\paragraph{Visual reasoning}
Early visual reasoning approaches used attention mechanisms to boost Visual Question Answering (VQA) performance \cite{zhou2021trar} or trained models directly to enhance visual reasoning skills \cite{wang2023visionllm}. More recent methods incorporate prompting strategies to better guide models toward relevant visual features \cite{zhang2024prompt}. For example, DDCoT \cite{zheng2023ddcot} breaks questions into sub-questions and uses external VQA models to generate rationales. Liu et al. \cite{liu2024enhancing} proposed a closed-loop framework that combines imagination and single-step reasoning, allowing models to iteratively refine their answers without further training.

Beyond reasoning strategies, some studies have enhanced understanding by explicitly manipulating visual inputs. For instance, Jiang et al. \cite{jiang2024joint} and Lin et al. \cite{lin2024draw} used bounding boxes during inference or training to help models focus on relevant objects, improving reasoning performance. Shao et al. \cite{shao2024visual} extended this by jointly processing raw and box-annotated images to guide and strengthen the reasoning process. Other approaches, such as Image-of-Thought prompting \cite{zhou2024image}, enable models to extract and generate both textual and visual rationales. Building on this, Zhang et al. \cite{zhang2024cocot} introduced a method for reasoning over multiple images by comparing visual similarities and differences.

Despite these contributions, we argue that, current approaches still rely heavily on textual generation for reasoning, with a shallow integration of visual information. Most visual manipulations, such as bounding boxes or segmentation, depend on external, pre-defined tools. As a result, the VLM itself lacks the ability to freely manipulate visual inputs, limiting its adaptability at inference time and preventing test-time scalability, as seen in text-only LLMs \cite{snell2024scalingllmtesttimecompute}. Moreover, these tool-based interventions break the end-to-end gradient flow, complicating the training and refinement of cross-modal reasoning capabilities. Instead, we propose a method that lets the LLM define manipulations independently by connecting its outputs directly to the vision encoder through a simple Multi-Layer Perceptron (MLP) module.


\section{Methodology}

We propose an approach to allow VLMs to do \textit{cross-modal} reasoning by introducing a lightweight visual reasoning module that connects the language understanding component to the visual encoding process. 
We release our training and evaluation code at \url{https://github.com/alessioGalatolo/VLM-Reasoning-for-Robotics}.

\subsection{Architecture choice}

Starting with Flamingo \cite{flamingo}, the prevailing approach for (open) VLMs has been to combine a pretrained vision encoder (e.g., CLIP \cite{clip}) with a pretrained or fine-tuned LLM through alignment training \cite{li2025benchmark}. Our method is specifically designed for such architectures and is compatible with widely used models, including
LLaVA-OneVision \cite{llava1.5}, Qwen 2.5 VL \cite{Qwen2.5-VL} and Gemma 3 \cite{gemmateam2025gemma3technicalreport}. Moreover, due to its minimal requirements, our approach can potentially be extended to a broader range of architectures beyond these examples.

\subsection{Visual reasoning module}

This module is attached to the final layer of the language model, receives its hidden representation and connects it to the vision encoder, effectively forming a reasoning loop between the two architectures. More specifically, given an input sequence of a prompt (i.e., a \textit{query}) and an image, we take the hidden states relative to the image, after it was processed by the language model alongside textual information. The visual reasoner then projects that back into the input space of the encoder, adding it to the image. By adding reasoning information before the image is fed into the encoder, we enable the model to reinterpret visual content in light of textual context and reasoning. At the end of this process, a second forward pass is done with the original prompt and image, plus the newly encoded one. Figure \ref{fig:complete_system} illustrates in detail this process. 

For the architecture of the visual reasoning module, we adopt a gated MLP: 
\begin{equation*}
    \sigma(W_g x) \odot W_p \left(\text{Dropout}\left(W_2 \cdot \text{GELU}(W_1 x)\right)\right)
\end{equation*}
The MLP's input and output dimensions match the hidden dimension of the LLMs; while the MLP's hidden dimension is set to double of the input/output dimension.

The Visual reasoner also comprises of a `patch unmerger', used to project back from the LLM representation space into the number of patches expected by the encoder.

\subsection{Training strategy}

We illustrate our training procedure in Algorithm \ref{alg:training}.

\begin{algorithm}[hb]
\caption{Detailed training procedure}
\label{alg:training}
\begin{algorithmic}[1]
\Require Dataset \( \mathcal{D} \), language model \( f_{\mathrm{LM}} \), visual reasoner \( \mathbf{r} \), visual encoder \(f_{\mathrm{VE}}\)

\State Inject LoRA adapters into \( f_{\mathrm{LM}} \)
\For{image \(x_v\), text \(x_q\) \( \in \mathcal{D} \)}
    \State \textbf{First forward pass (LoRA enabled if available):}
    \
    \State \quad \(H_v \gets f_\mathrm{VE}(x_v)\) 
    \State \quad Compute hidden states \( H \gets f_{\mathrm{LM}}(x_q, H_v) \)
    \State \quad Extract visual hint \( z \gets H_{\text{last}} \)
    \State \quad Set image reasoning: \( \mathbf{r}(z) \)

    \State \textbf{Second forward pass (LoRA disabled always):}
    \State \quad \(H_{\bar{v}} \gets f_\mathrm{VE}(x_v + \mathbf{r}(z))\) 
    \State \quad Compute prediction \( \hat{y} \gets f_{\mathrm{LM}}(x_q, H_v, H_{\bar{v}}) \)
    \State \quad Compute loss \( \mathcal{L}(\hat{y}, \text{labels}) \)

    \State Update parameters of \( \mathbf{r} \) and LoRA if available
\EndFor
\end{algorithmic}
\end{algorithm}

\begin{figure*}[ht]
    \centering
    \includegraphics[width=1\linewidth]{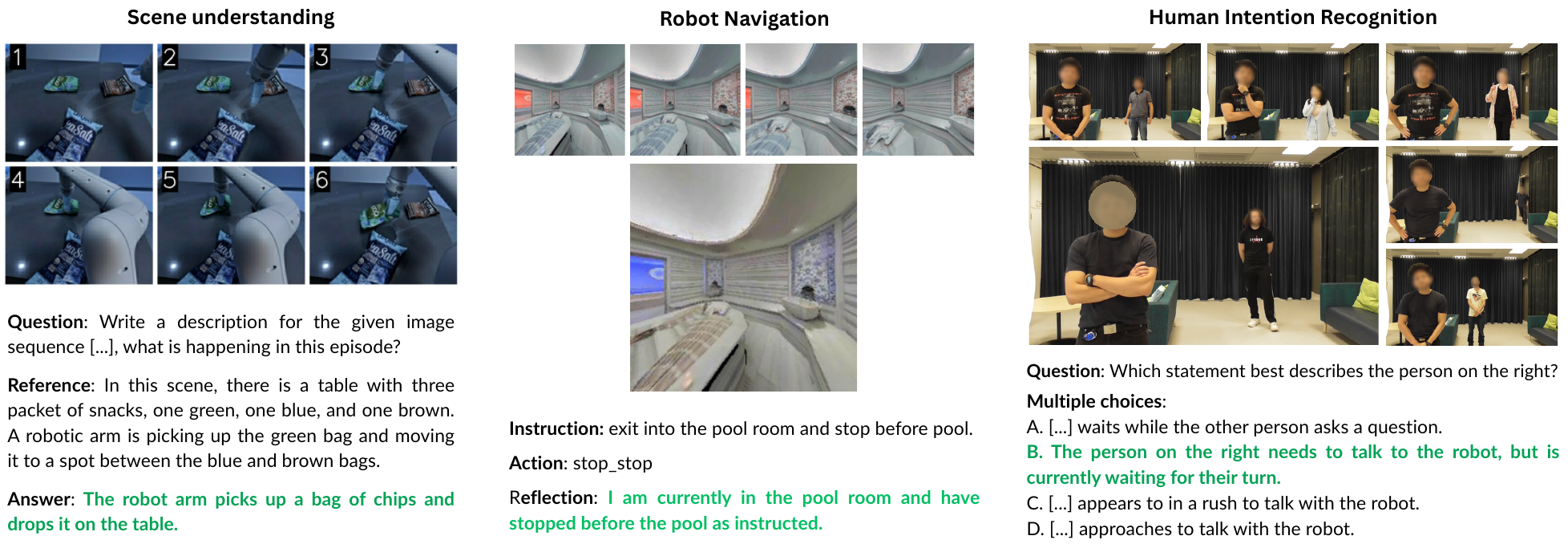}
    \caption{Examples of the datasets use for evaluation. The Mementos-Robotics dataset \cite{wang2024mementos} provides sequential images with scene descriptions; the Navigation benchmark \cite{puig2023habitat3} contains robot trajectories in simulation; and our human-intention recognition dataset captures interactions between humans and a social robot.} 
    \label{fig:datasets}
\end{figure*}

During training, we perform two passes for each step. In the first, the model is given the user query and the input image and performs a standard forward pass. From here, we take the final hidden state of the LLM, for the tokens corresponding to the image. We then process these representations through the visual reasoning module and the `unmerger'. The result is then added to the original image after embedding, but before it is encoded. This produces a new encoding of the image that takes into account the feedback from the LLM.

In the second pass, we feed the user query, the original image, and the new image to the model. The loss is then computed (only) from this pass and is used for backpropagation. The loss is thus back-propagated from the end of the language model, to the vision encoder, to the visual reasoner---updating its weights while keeping the LLM and vision encoder \textit{frozen}.

In our experiments, we also test the integration of LoRA layers \cite{hu2022lora} in the language model. We only enable these layers for the first pass, to aid the model in providing feedback to the vision encoder, and keep it disabled otherwise. The total training parameters amount to less than 1.7\% of the original model and less than 3\% when also including LoRA.

\subsection{Training data}
For training, we use the Visual-CoT dataset \cite{shao2024visual}, which provides image–question inputs paired with a reasoning output. We intentionally selected a non-specialized dataset. While training on a robotics-specific dataset would likely yield higher gains, it would also bias the evaluation against the base model (which was not trained on such data) and risk narrowing the applicability of VLMs. Using a general dataset instead preserves a key advantage of VLMs, i.e., their ability to address a wide range of problems beyond domain-specific settings. 
To reduce computational demand for training, we resize all images to 360p resolution.

\subsection{Inference}
Suppose input $<$image, question$>$, an initial forward pass is done (with LoRA enabled if available) to generate the visual reasoning hint $z$. The output of the visual reasoner $r(z)$ is used to change embedding of the vision encoder. The LLMs is then passed both the old and the new image and it generates its reply through standard practices (see Figure \ref{fig:complete_system}). 

\subsection{Optimization and infrastructure}

Optimisation was performed using the AdamW optimiser, with the model checkpointed at regular intervals. Training was conducted on 4$\times$A100 (40GB), completing each stage in a full epoch using half-precision (bf16). We sweep over 7 learning rates $\eta \in [1e-2, 1e-5]$, equally spaced. We also experiment with placing the image before and after the prompt (details in Section \ref{sec:ablations}). At the end, we select the top 2 configurations, using the MME-CoT as validation dataset \cite{jiang2025mmecot}, and merge them via linear interpolation \cite{ilharco2023editing}. Training lasted between 1h30 and 2h45, depending on configuration and model size.

\section{Evaluation}
We evaluate our approach against state-of-the-art VLM baselines and conduct several ablation studies. For the baselines, we consider three models: Qwen 2.5 VL 7B\cite{Qwen2.5-VL}, Gemma 3 4B \cite{gemmateam2025gemma3technicalreport} and LLaVA OneVision 1.5 4B\cite{llava1.5}. We then test the performance against three robotic-centred benchmarks: (i) robot navigation (ii) scene understanding (iii) human intention recognition. We use a greedy decoding strategy with no sampling for all the benchmarks. 
\begin{itemize}
    \item \textbf{Robot navigation}: We follow the pipeline from \cite{duan2025navigation}, where a robot navigates in the Habitat simulation \cite{puig2023habitat3} towards a goal specified in natural language. Robot actions are parsed from text (JSON) and can be: stay, move forward (0.25 step size), rotate left/right (15 degrees).
    \item \textbf{Scene understanding}: We use the Mementos-Robotics benchmark \cite{wang2024mementos}, which provides sequential images annotated with scene descriptions. 
    \item \textbf{Human intention recognition}: We collected our own dataset, described in the following section.
\end{itemize}

\subsection{Human intention recognition dataset}
\label{sec:CustomDataset}

Since one of our main goals is to analyse human behaviour, with a particular focus on recognising human intentions, we require data with behaviour-specific annotations and a perspective resembling a robot's point of view. Existing robotics datasets often focus on pose prediction \cite{interact} or lack publicly available code/data \cite{socialnav}. Datasets such as UE-HRI \cite{BenYoussef2017} and JPL First-Person Interaction \cite{ryoo2013first} meet some of these requirements, but they are limited in recording quality and behaviour variability. To overcome these limitations, we construct and annotate a new dataset\footnote{Our application for ethical approval to the local review authority concluded that approval was not required, as the study does not fall under the provisions of the National Ethical Review Act.}. 

Our dataset consists of audio-video recordings collected during a human-robot interaction study with 10 participants (3 female, 6 male, and 1 who preferred not to disclose gender). Participants had an average age of 30.9 years ($SD=4.77$),  most reported little prior experience with robots ($M=2.00$, $SD=1.33$ on a [1–5] scale), and all but one worked in engineering or technology-related fields.

The study featured the \href{https://www.furhatrobotics.com/}{Furhat} social robot, acting as a tourism assistant at an information desk. A \textit{confederate}, playing the role of a tourist, initiated a conversation with the robot before participants entered the recording area. After signing a consent form, participants were then instructed to request information from the robot under three urgency conditions: \textit{not rushed}, \textit{somewhat rushed}, and \textit{very rushed}. The duration of the ongoing interaction was shortened accordingly to naturally induce different levels of time pressure. Because conversational intervention can take multiple forms 
, the study was designed to observe whether and how participants chose to intervene depending on the urgency condition.

Our goal is to clearly specify participants' intentions while allowing them to express these intentions in a natural behaviour. This approach grounds their intentions and captures rich audiovisual data to study interaction strategies.

\subsubsection{Annotation and benchmark adaptation}

We annotated five types of participant behaviours from the video recordings: (1) waiting for their turn to speak with the robot, (2) approaching to interrupt the conversation, (3) calmly signalling intent to speak, (4) urgently signalling intent to speak, and (5) interacting with the robot while the confederate waited. In total, 188 events were annotated.

We pre-processed this set by converting each annotated caption into two multiple-choice questions, one referring to the inactive person (already engaged with the robot) and one to the intervening person (attempting to interact). First, we normalised captions by replacing explicit positional markers (`left'/`right' person) with a placeholder to build a pool of candidate templates. In a second pass, we restored the appropriate position for each instance and constructed four-option multiple-choice questions, ensuring that the correct description was always included alongside randomly sampled distractors from the option pool. Images were resized to a standard resolution (360p), and each question–answer pair was stored together with the corresponding processed image. The final dataset contains 376 samples, where each sample has a single associated frame and one question with four possible answers; the samples do not contain any textual information about the scene.

We do not make any splits of this dataset, as it is \textit{only} used for the final evaluation.

\subsection{Ablations and variations}
\label{sec:ablations}
To better understand the contributions of different components to the performance of our method, we conduct several ablation experiments. First, we remove the input image for the second embedding; in this scenario, the model receives the original image in the first pass and a completely artificial image (made by the LLM itself) in the second pass. Second, we remove the visual reasoning module, using only the hidden state from the LLM and project it back into the encoder's space (through the \textit{unmerger}). Third, we experiment with ablating a stage, i.e., training only stage 1 or only stage 2. Finally, we test two variations in the order of inputs, passing the image either \textit{before} or \textit{after} the prompt (i.e., user query). Intuitively, passing the image \textit{before} the prompt prevents its processed hidden states to take into account the prompt due to the causal masking used in LLMs.

\subsection{Metrics}
To provide a comprehensive evaluation of our method, we use a variety of metrics across benchmarks. For the navigation benchmark, we report the final distance to the goal, averaged across episodes. For the Mementos benchmark, which requires open-ended scene descriptions, we adopt an LLM-as-a-judge approach, scoring each generation from 1 to 5 based on its overlap with the ground-truth. For our intention recognition benchmark we opted for multi-choice questions and thus report accuracy as the evaluation metric. 

\section{Results}

Table~\ref{tab:main_results} reports the performance of our method compared to plain model baselines on all three tasks. On the Qwen 7B backbone, our approach achieves consistent improvements across all metrics: the final distance to goal in navigation is reduced (from 7.787 to \textbf{7.530}), open-ended description scores on Mementos increase (from 2.261 to \textbf{2.318}), and accuracy on intent recognition rises (from 34.04\% to \textbf{36.97\%}). For Gemma 4B, improvements are less uniform: our method provides a substantial boost in Mementos score (1.693 to \textit{1.804}) and intention accuracy (20.84\% to \textit{31.65\%}), but navigation distance slightly worsens (7.977 to 8.014). A similar pattern emerges for LLaVA 4B: our method improves Mementos score (from 2.201 to \textit{2.256}) and intention accuracy (from 20.74\% to \textit{25.53\%}), while navigation distance again slightly worsens (from 7.832 to 8.114).

Taken together, the results show that the proposed visual reasoning module yields the largest relative gains on scene description (Mementos) and human intention recognition, especially for models with lower initial performance. 

The improvements are less significant (or sometimes negative) for the navigation task on Gemma and LLaVA. Here, we do a manual analysis of the generations, and for Gemma, we reduce this to a general difficulty of the model (both plain Gemma and our approach) in producing properly formatted output (i.e., JSON with action to take), resulting in the agent often skipping a `step' due to the missing or incorrectly formulated action. On the case of LLaVA, we notice a very frequent refusal behaviour from the model (which does not happen in the other tasks). The Qwen backbone consistently benefits from our method, with gains in all three domains.

\begin{table}[]
    \centering
    \adjustbox{width=\columnwidth}{
    \begin{tabular}{@{} lccc @{}}
            \toprule
            &  \textbf{\textit{Navigation}} &\textbf{\textit{Mementos}} & \textbf{\textit{Intentions}}\\
            \cmidrule(lr){2-4}
            \textbf{Model} & \textbf{Distance} $\downarrow$ & \textbf{Score} $\uparrow$ & \textbf{Accuracy} $\uparrow$  \\
            \midrule
            Plain Gemma 4B & \textit{7.977} & 1.693 & 20.84\%\\
            Ours (Gemma 4B) & 8.014 & \textit{1.804} & \textit{31.65\%}\\
            \midrule
            Plain LLaVA 4B & \textit{7.832} & 2.201 & 20.74\%\\
            Ours (LLaVA 4B) & 8.114  & \textit{2.256} & \textit{25.53}\%\\            
            \midrule
            Plain Qwen 7B & 7.787 & 2.261 & 34.04\%\\
            Ours (Qwen 7B) & \textbf{7.530}  & \textbf{2.318} & \textbf{36.97\%} \\
            \bottomrule

    \end{tabular}
    }
    \caption{Performance of our method compared to the plain model. We highlight in \textbf{bold} the best performing model in each column and in \textit{italic} the best version of each family.}
    \label{tab:main_results}
\end{table}

\subsection{Ablation study}

\begin{table}[]
    \centering
    \adjustbox{width=\columnwidth}{
    \begin{tabular}{@{} lccc @{}}
        \toprule
        &  \textbf{\textit{Navigation}} &\textbf{\textit{Mementos}} & \textbf{\textit{Intentions}}\\
        \cmidrule(lr){2-4}
        \textbf{Variation} & \textbf{Distance} $\downarrow$ & \textbf{Score} $\uparrow$ & \textbf{Accuracy} $\uparrow$  \\
        \midrule
        Removing original & 7.764 & 1.950 & 34.31\%\\
        No MLP & 7.831 & 1.980 & \textbf{37.50}\%\\
        Image first & \textbf{7.685} & \textbf{2.000} & 28.46\%\\
        Prompt first & 8.056 & 1.744 & 25.53\%\\
        \bottomrule
    \end{tabular}
    }
    \caption{Ablation studies using Qwen only. \textbf{Bold} highlights the best performing setting in each column.}
    \label{tab:ablations}
\end{table}

To isolate the contributions of each component, Table~\ref{tab:ablations} presents results of several ablations using the Qwen backbone. Removing the original image in the second pass or ablating the MLP from the visual reasoner both lead to marked drops in performance on all benchmarks except intention recognition where the accuracy is still competitive at 34.31\% without the original image and 37.50\% without the MLP, compared to 36.97\% with the full method. We attribute this effect to the `patch unmerger' which is still a learned component (as it is strictly necessary for the method) who could partially compensate for the absence of the MLP. The navigation and Mementos metrics degrade in both ablations.

Ablating the order of input modalities (passing the image before or after the prompt) highlights how our initial assumptions were mistaken. The ``image first'' variant achieves 28.46\% accuracy on intentions, while ``prompt first'' fares lower at 25.53\%. Both are far below the complete method. The ``prompt first'' variant also scores lower on both the other two benchmarks.

In summary, the results confirm that both the use of the original image in the second pass and the presence of the MLP-based visual reasoner are necessary for best performance. The order in which the input modalities are fed into the LLM also matters, with ``image before prompt'' outperforming other settings.

\subsection{Baselines}
For our method, we experiment with placing the image before or after the user query, expecting increased performance for the latter (due to causal masking). Unexpectedly, our experiments reveal that our method performs best when the image appears before the user query. We attribute this issue to how the VLMs were initially trained, preferring one particular structure over the other. We empirically confirm this on the \textit{base} models and report up to $30\%$ performance degradation when swapping image and user query. 

We further test another variation in the baselines. Here, we want to establish whether the improved performance of our method could be due to using two images instead of one. We thus provide our baselines two times the original image, in the same template as our method. This test yields even greater performance degradation, further supporting the effectiveness of our method. 

Table \ref{tab:main_results} reports only the best results for the baselines.

\subsection{Resource Consumption}

\begin{table}[]
    \centering
    \adjustbox{width=\columnwidth}{
    \begin{tabular}{@{} lccc @{}}
            \toprule\\
            \textbf{Model} & \textbf{Avg TFLOPs} $\downarrow$ & \textbf{Samples/sec} $\uparrow$ & \textbf{Peak memory (GB)} $\downarrow$  \\
            \midrule
            Baseline & 7.06 & 4.24 & 15.9 \\
            Ours & 20.39 & 1.27 & 16.32\\
            \bottomrule
    \end{tabular}
    }
    \caption{Resource consumption of our method compared to baseline. We report the average numbers when evaluating the Qwen model on our intentions dataset, tested on consumer hardware: NVIDIA RTX 3090.}
    \label{tab:resources}
\end{table}

Table \ref{tab:resources} reports the computational overhead introduced by our method relative to the baseline, measured on an NVIDIA RTX 3090 during evaluation of the Qwen model on the human-intention recognition dataset. The additional cost is a direct consequence of the dual forward-pass design: performing two passes through both the vision encoder and the language model roughly triples the average TFLOPs (from 7.06 to 20.39) and reduces throughput from 4.24 to 1.27 samples per second. Despite this increase in compute, the memory footprint remains modest, rising by less than 3\% (from 15.9 to 16.32 GB), which is consistent with the lightweight nature of the visual reasoning module and the fact that the additional parameters account for less than 3\% of the original model. The increased latency is therefore primarily attributable to the extra inference pass rather than to any substantial growth in model size. Crucially, the method remains deployable on a single consumer-grade GPU, and a throughput of over one sample per second is sufficient for real-time deployment in robotics applications such as human-intention recognition and navigation, where perception typically operates at low frequencies. For scenarios demanding higher throughput, further optimisation through quantisation or hardware acceleration could be readily applied without altering the method itself.

\section{Discussion}

Our results provide empirical evidence that introducing a lightweight visual reasoning module improves cross-modal reasoning across multiple robotics-centred tasks. This improvement is consistent across three different families of vision-language models, and particularly substantial for open-ended tasks like scene description and human intention recognition. In this section, we discuss both task-specific observations and broader implications of our design choices.

\subsection{Interpretable cross-modal feedback loops}

The core contribution of our method is the explicit feedback loop from the language model to the vision encoder. This loop, instantiated through a compact MLP and unmerging mechanism, enables the language model to modulate visual processing based on its understanding of the task and input prompt. Unlike most existing VLMs, where vision is passively embedded once and never updated, our approach reuses the language output to alter the visual embedding. Notably, our method does not require backpropagation through the vision encoder nor modifications to the base models, making it practical to integrate into existing VLM pipelines.

Further, our method provides a seamless integration of the two modalities, without breaking the gradient flow. In this initial work, we opted to use a general-purpose dataset (i.e., Visual-CoT \cite{shao2024visual}) to avoid any kind of cross-contamination and to show how our method is fit to improve a broader class of problems. However, it can be easily trained on more specialised data, potentially yielding far greater performance enhancement.

\subsection{Performance across tasks}

The three evaluated robotics-centred tasks differ in modality balance: navigation relies more heavily on structured outputs and spatial understanding; scene description requires contextual visual parsing; and intention recognition involves subtle social cues and multi-party reasoning. Our method yields consistent improvements in the latter two tasks, suggesting it is particularly effective when high-level visual semantics are critical.

For navigation, however, the gains are less consistent. In particular, performance with the Gemma and LLaVA backbone slightly declines, which we attribute not to failures in visual reasoning but to limitations in the model's ability to reply with well-structured output. 

\subsection{Design trade-offs and methodological lessons}

Despite its simplicity, our method yields measurable gains with some trade-offs:

\subsubsection{Image reuse in forward passes} We show that using the original image in both passes is essential---ablating this reduces performance across all tasks. This suggests that visual reinterpretation, rather than visual replacement, is a more stable strategy.
    
\subsubsection{Ordering of modalities} Although our design places the image after the prompt to exploit causal masking, performance unexpectedly degrades for models initially trained with the reverse ordering. 

\subsubsection{Baselines using image duplication} We tested whether performance gains could be attributed to simply providing more visual input (e.g., duplicating the same image). These baselines performed worse, supporting the necessity of the modulated second image as opposed to brute-force repetition.

These findings indicate that introducing cross-modal feedback is not just a matter of adding more data or capacity, but of strategically closing the loop between perception and interpretation.

\section{Conclusion}

We begin our work by investigating common challenges for robots situated in human-shared environments. Next, we target VLMs as general-purpose models, flexible in task definition and modality integration. Here, we introduce a novel approach for enhancing cross-modal reasoning in vision-language models through a lightweight visual reasoning module. Our intuition stands behind the idea that the `default' encoding of the visual input is not aimed at grasping specific small cues that may be required by the context (e.g., predicting upcoming human behaviour), while the language model part is in principle capable of looking for such cues. Our method enables a feedback loop from language interpretation to visual perception by injecting dynamically generated reasoning hints into the vision encoder. Across three robotics tasks---navigation, scene understanding, and human intention recognition---our method consistently outperforms strong baselines with minimal additional parameters. More broadly, our results challenge the dominant feedforward paradigm in vision-language integration. By demonstrating that even frozen (encoder) models can benefit from guided visual reinterpretation, we provide a new tool for building more adaptive and context-aware robotic agents.

Our work highlights the value of architectural asymmetry and feedback in multimodal models---a principle well-known in embodied cognition but rarely realised in VLMs.


\section*{ACKNOWLEDGMENT}
We thank Zixuan He for her work on implementing and debugging the LLaVA variant of our method.

The computations and data handling were enabled by resources provided by the National Academic Infrastructure for Supercomputing in Sweden (NAISS) at Alvis, C3SE (Chalmers) partially funded by the Swedish Research Council through grant agreement no. 2022-06725. This research was supported by the Horizon Europe EIC project \href{https://symaware.eu}{SymAware} under the Grant Agreement No. 101070802.

\bibliographystyle{IEEEtran}
\bibliography{references}

\balance

\end{document}